\documentclass[11pt]{article}

\usepackage[final]{acl}

\usepackage{times}
\usepackage{latexsym}

\usepackage[T1]{fontenc}

\usepackage[utf8]{inputenc}

\usepackage{microtype}

\usepackage{inconsolata}

\usepackage{graphicx}
\usepackage{subcaption}
\usepackage{amsmath}
\usepackage{booktabs}

\usepackage[most]{tcolorbox}
\usepackage{fvextra}
\usepackage{tabularx}
\usepackage{array}
\usepackage{fontawesome}
\usepackage{multirow}

\title{The ROI of Small Examples: Optimizing In-Context Learning Under Token Budgets}

\author{Anxo Perez \\
  IRLab, CITIC \\
  Universidade da Coruña \\
  A Coruña, Spain \\
  \texttt{anxo.pvila@udc.es} \\\And
  Jorge Gabín \\
  IRLab, CITIC \\
  Universidade da Coruña \\
  A Coruña, Spain \\
  \texttt{jorge.gabin@udc.es} \\\And
  Javier Parapar \\
  IRLab, CITIC \\
  Universidade da Coruña \\
  A Coruña, Spain \\
  \texttt{javier.parapar@udc.es} \\}

\begin{document}
\maketitle
\begin{abstract}
In-context learning (ICL) improves large language model performance at inference time without updating model parameters. However, these gains are not free: in practical deployments, longer prompts directly increase inference cost and latency. Under a fixed token budget, prompt construction therefore becomes a resource-allocation problem. While prior work has often focused on how many examples to include in ICL, we study the complementary question of whether the available budget is better spent on a few long demonstrations or on a larger number of shorter ones. We study this aspect in question answering (QA), by organizing candidate demonstrations into length-based bins and constructing prompts from examples within the same length range under fixed budgets. Experiments on two QA benchmarks (\textit{Beat the AI} and \textit{Natural Questions}), using six instruction-tuned LLMs, show that shorter demonstrations are generally the most effective use of prompt space. This effect is especially strong on \textit{Beat the AI}, and remains visible on NQ, although gains are smaller and saturate earlier in the long-context setting. An additional natural language inference experiment on ANLI R3 finds that one of the two shorter bins performs best for five of six models, although longer demonstrations remain competitive. Overall, our results highlight demonstration granularity as an important factor in prompt design.
\end{abstract}

\section{Introduction and Related Work}

In-context learning (ICL) has become an attractive way to adapt LLMs to downstream tasks without updating model parameters~\cite{brown2020language}. By conditioning the model on a set of demonstrations, ICL can improve performance at inference time in tasks such as Question Answering (QA)~\cite{li2023fewshot}. However, every additional demonstration increases prompt length, which in practical deployments directly affects inference cost and latency. As a result, many real-world applications must operate under fixed token budgets. This constraint turns prompt construction into a resource allocation problem. Given a limited number of tokens, should a practitioner include a few long demonstrations, or allocate the same budget to a larger number of shorter ones?

This trade-off is especially relevant in QA, where demonstration length is often driven by the amount of supporting context included with each example. A longer demonstration may expose the model to richer evidence and more complete question--answer mappings, but it also consumes budget that could otherwise be spent on additional examples. Conversely, shorter demonstrations may provide less contextual detail individually, yet allow the model to observe a broader range of answer patterns within the same prompt. As a result, prompt construction under budget constraints is not simply a matter of adding more demonstrations, but of deciding how to distribute limited context space between \emph{depth} per example and \emph{breadth} across examples.

Prior work has shown that ICL performance depends strongly on prompt construction~\cite{lu2022fantastically,zhao2024dynamicdemonstrationscontrollerincontext}. Some studies have explored how to select better demonstrations, including retrieval-based and budget-aware strategies that dynamically decide which examples to include or how many to use under context constraints \citep{Xu2023,Purohit2025,zhou2023efficientpromptingdynamicincontext}. Other work shows that demonstration organization also matters: example ordering can significantly affect ICL performance, and long-context studies suggest that additional context is not always exploited effectively, especially when relevant information appears in less favorable positions \citep{zou2025many,liu2023lost,bertsch2025context}. Despite these advances, much less is known about the direct trade-off between \emph{demonstration length} and \emph{demonstration count} under a fixed prompt budget. In particular, it remains unclear whether the same token budget is better spent on many concise demonstrations or on fewer but longer ones.

In this paper, we group candidate demonstrations into length-based bins and compare prompting strategies under controlled token budgets, varying the number and size of examples within a shared budget ceiling. Our main study covers two QA benchmarks: \textit{Beat the AI}~\cite{bartolo2020beat}, which provides a challenging adversarial setting, and a subset of Natural Questions (NQ), which offers a complementary QA benchmark with much longer contexts. To test whether the effect generalizes across architectures and scales, we evaluate six instruction-tuned LLMs spanning three model families and two sizes per family. We further explore its applicability beyond extractive QA through an additional experiment on ANLI R3, a natural language inference benchmark~\cite{nie-etal-2020-adversarial}.

Our results show that, under a fixed token budget, shorter demonstrations are generally the most effective use of prompt space in the QA settings studied. This effect is especially strong on \textit{Beat the AI}, where the shortest bin yields the best results across all evaluated models. In NQ, the gains are smaller and saturate earlier in the long-context setting. On ANLI R3, one of the two shorter bins yields the best observed result for five of six models, while longer demonstrations remain competitive, indicating that the preferred allocation depends on the task and model. These findings provide practical guidance for budget-constrained ICL and highlight demonstration granularity as an important and underexplored dimension of prompt design\footnote{\faicon{github}: \scriptsize \url{https://github.com/IRLab-UDC/roi-small-samples-in-icl}}.
\section{Methodology}

We study prompt construction for QA-oriented ICL under a fixed token budget. Our goal is to compare two competing uses of the same budget: allocating it to a smaller number of longer demonstrations, or to a larger number of shorter ones. We address this setup as a controlled token-allocation problem in which prompts are matched by total token length while the average size of the demonstrations varies.

Let a prompt be defined as $P=(I,D,t)$, where $I$ is the task instruction, $D=\{d_1,\dots,d_n\}$ is a set of in-context demonstrations from the training set, and $t=(x_q,c_q)$ is the test instance consisting of a question $x_q$ and its associated context $c_q$. Each demonstration $d_i=(x_i,c_i,y_i)$ is represented as a full multi-turn example containing a question, context, and answer. Its length, denoted $L(d_i)$, is the total tokenized length of the complete formatted demonstration. The total prompt length is therefore
\[
L(P)=L(I)+L(t)+\sum_{i=1}^{n}L(d_i),
\]
and must satisfy the budget constraint $L(P)\leq T$, where $T$ is the maximum token budget for the full prompt.

We partition the pool of candidate demonstrations into length-based bins using the Freedman--Diaconis rule\footnote{Full set of bins obtained for all datasets, with the corresponding token-length histograms, is provided in Appendix~\ref{app:bins}.}, which yields data-driven ranges of similarly sized examples. For \textit{Beat the AI}, this results in four bins: $B_1=[48,84]$, $B_2=[195,231]$, $B_3=[342,377]$, and $B_4=[452,487]$. The same procedure is applied independently to the Natural Questions subset, yielding four larger bins: $B_1=[570,1434]$, $B_2=[2300,3163]$, $B_3=[4029,4892]$, and $B_4=[6622,7486]$.

For each budget level $T$ and each length bin $B_j$, we construct prompts using only demonstrations sampled at random from $B_j$, so that all examples in a prompt come from the same length range. Starting from an empty prompt, we iteratively add demonstrations until including another one would exceed $T$. Because shorter bins contain more compact examples, they typically allow more demonstrations to fit within the same budget, whereas longer bins result in fewer included examples. The budget is applied to the entire prompt, including the task instruction, all demonstrations, and the final test instance. Since the content distribution across bins is comparable and the sampled examples do not differ substantially in difficulty, this setup isolates the trade-off of interest: whether, under the same token expenditure, LLMs benefit more from broader example coverage or from richer individual demonstrations.

\section{Experiments}

\begin{figure*}[t]
    \centering

    \captionsetup[subfigure]{skip=2pt}
    
    \begin{subfigure}[t]{0.49\textwidth}
        \centering
        \includegraphics[width=\linewidth]{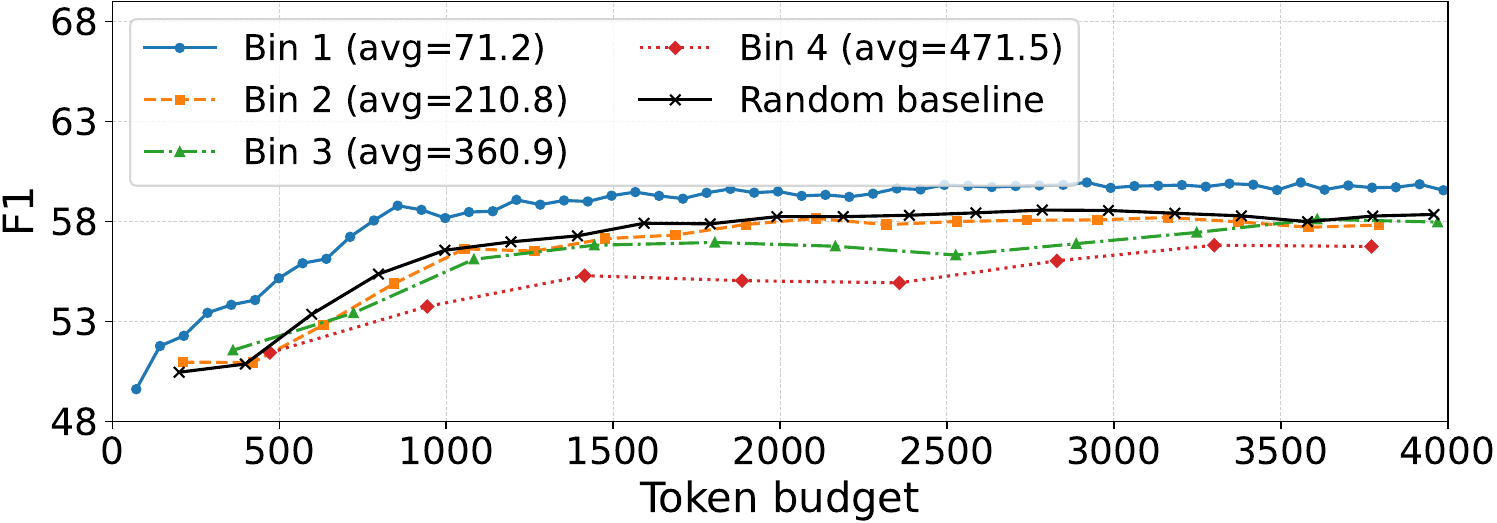}
        \caption{Gemma 4B}
        \label{fig:gemma4b}
    \end{subfigure}\hfill
    \begin{subfigure}[t]{0.49\textwidth}
        \centering
        \includegraphics[width=\linewidth]{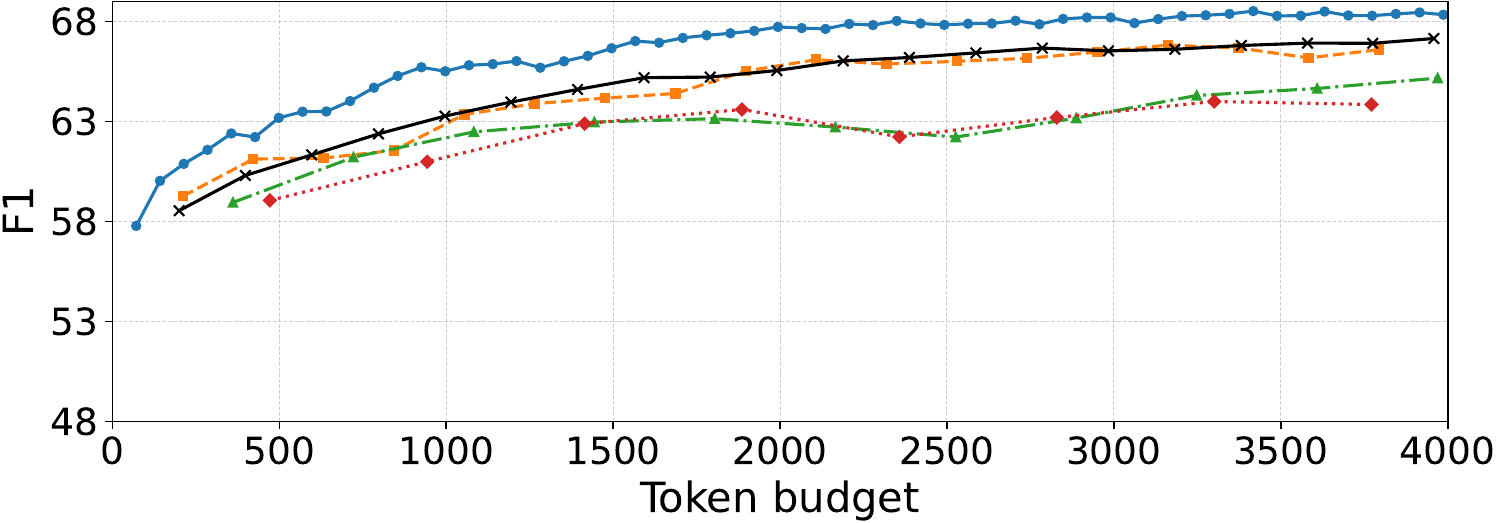}
        \caption{Gemma 27B}
        \label{fig:gemma27b}
    \end{subfigure}

    \vspace{-0.2em}

    \begin{subfigure}[t]{0.49\textwidth}
        \centering
        \includegraphics[width=\linewidth]{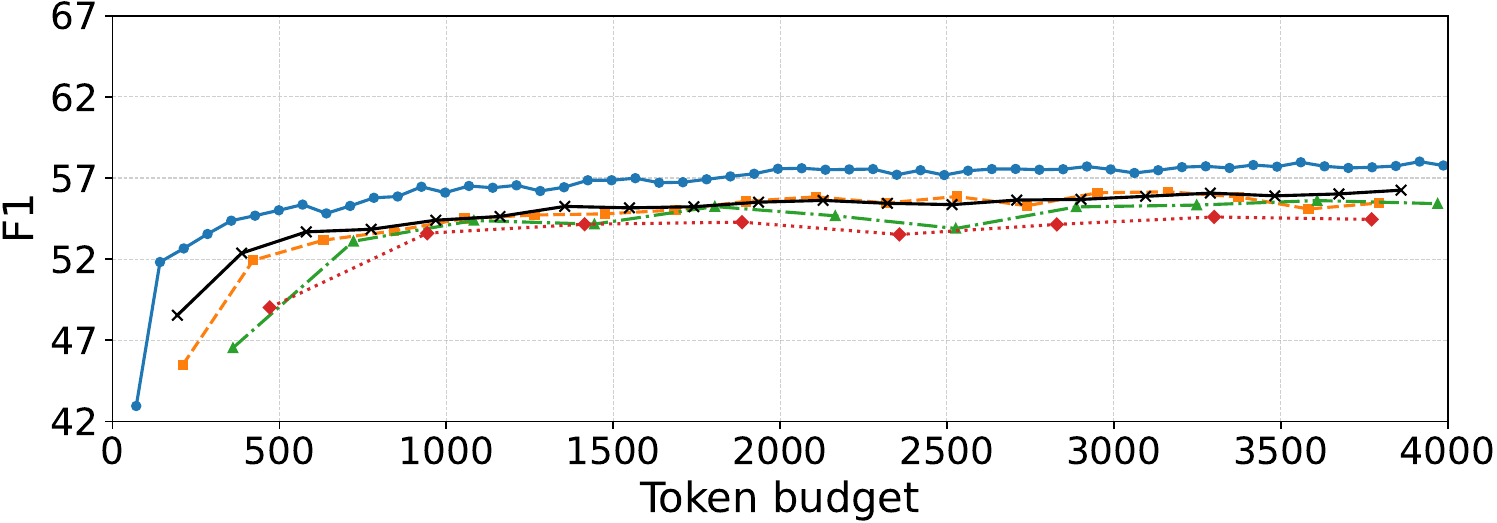}
        \caption{Phi 4B}
        \label{fig:phi4b}
    \end{subfigure}\hfill
    \begin{subfigure}[t]{0.49\textwidth}
        \centering
        \includegraphics[width=\linewidth]{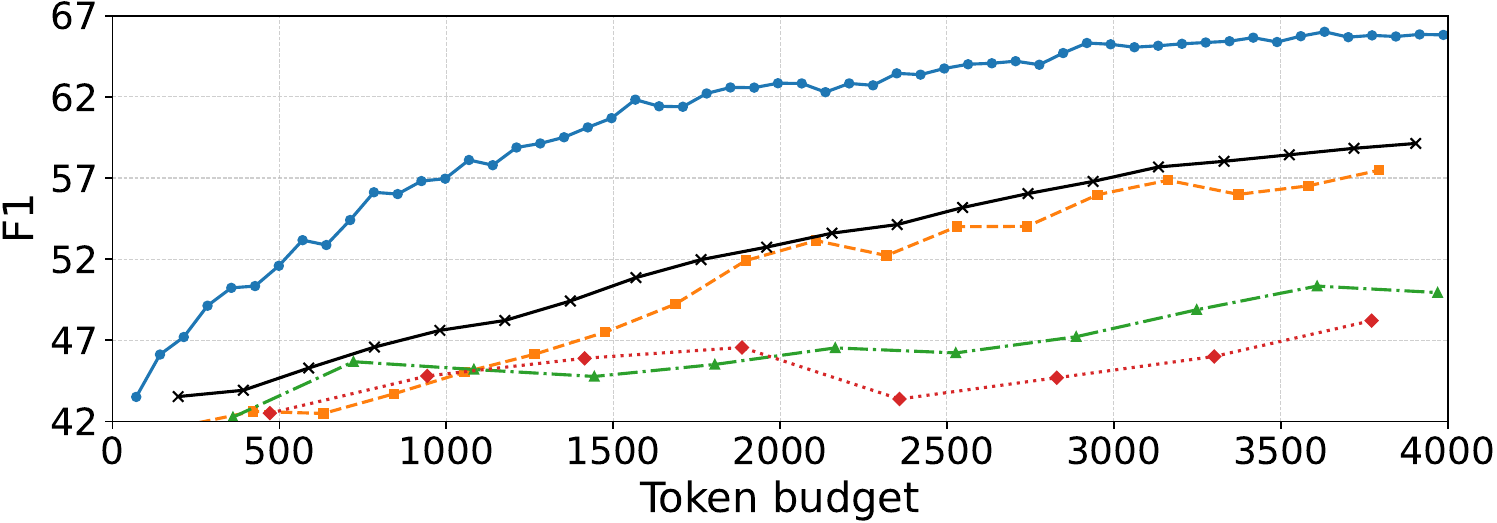}
        \caption{Phi 14B}
        \label{fig:phi14b}
    \end{subfigure}

    \vspace{-0.2em}

    \begin{subfigure}[t]{0.49\textwidth}
        \centering
        \includegraphics[width=\linewidth]{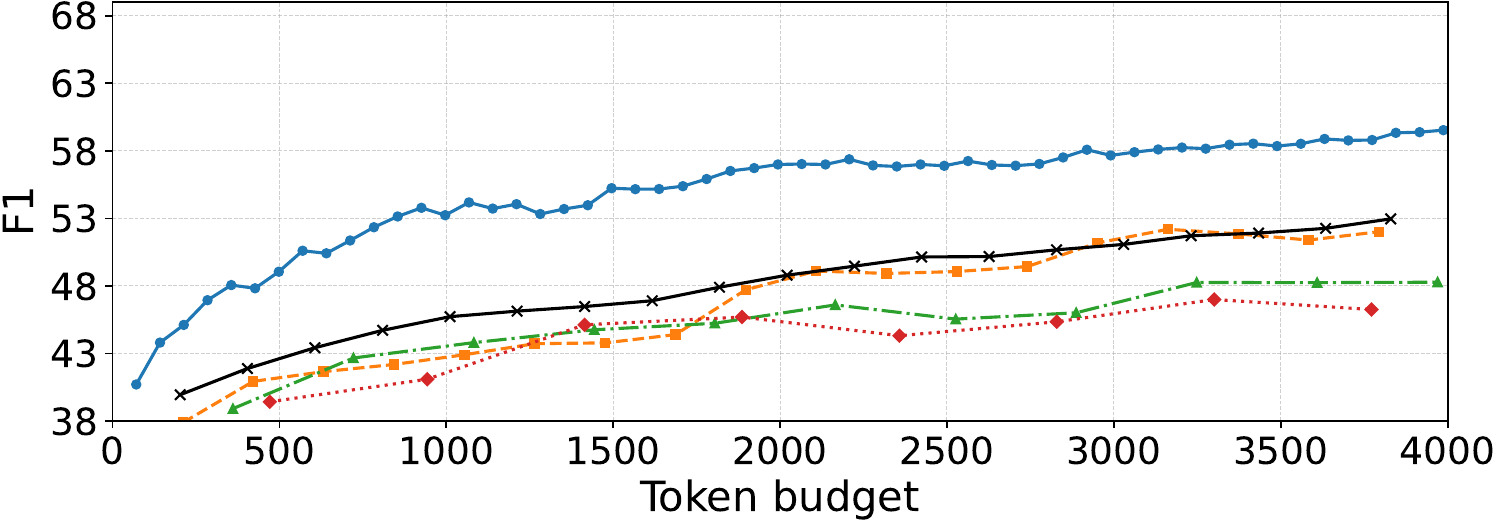}
        \caption{Qwen 4B}
        \label{fig:qwen4b}
    \end{subfigure}\hfill
    \begin{subfigure}[t]{0.49\textwidth}
        \centering
        \includegraphics[width=\linewidth]{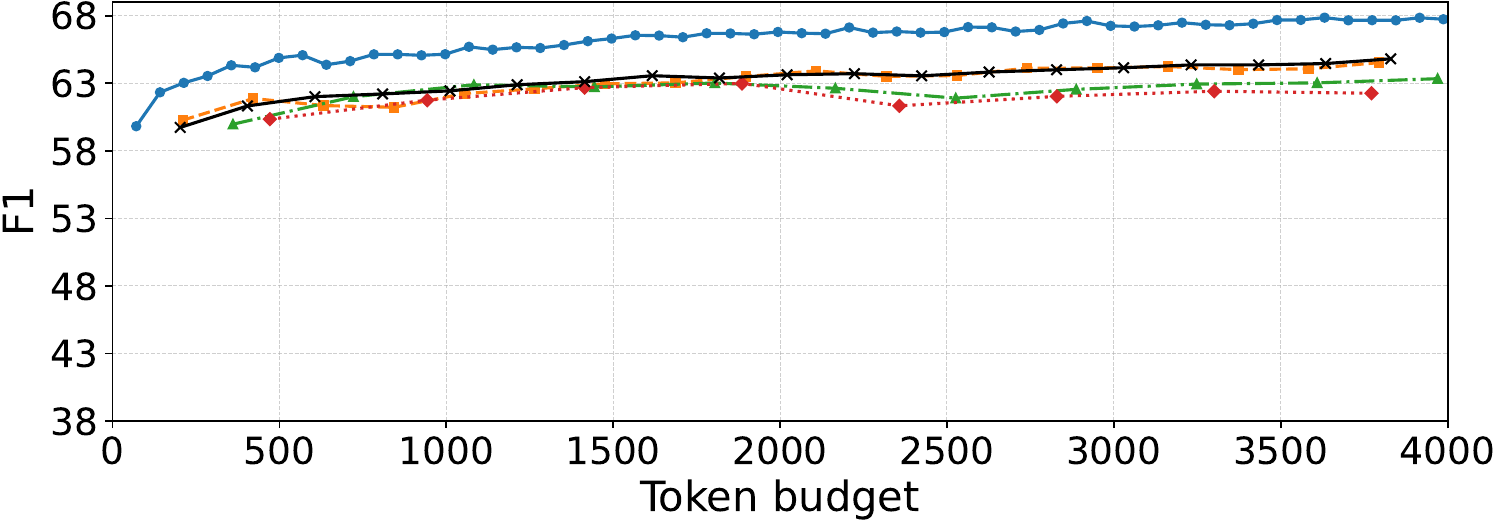}
        \caption{Qwen 30B}
        \label{fig:qwen30b}
    \end{subfigure}

    \vspace{-0.4em}
\caption{Average F1 on \textit{Beat the AI} as a function of prompt budget for the six evaluated LLMs. Each curve corresponds to a length bin, and random baseline samples demonstrations from the full pool without bin constraints.}    \label{fig:model_size_comparison}
\end{figure*}

We evaluate our approach on two QA benchmarks. The first is \textit{Beat the AI}~\cite{bartolo2020beat}, an adversarial reading comprehension benchmark, where questions were collected to deliberately challenge existing QA systems. The second is a subset of Natural Questions (NQ)~\cite{kwiatkowski-etal-2019-natural}, included to assess whether the observed trends extend beyond the adversarial setting. For NQ, we restrict the candidate training pool to examples shorter than 7{,}500 tokens for computational feasibility, and evaluate on a subset of 500 test queries.

We experiment with six instruction-tuned LLMs from three families and two model sizes per family: Qwen 3 (4B and 30B), Phi (Phi-4-mini and Phi-4), and Gemma 3 (4B and 27B). For each model, we compare zero-shot prompting against in-context prompting under fixed token budgets, using prompts built from a single length bin at a time\footnote{Full prompt format is shown in Appendix~\ref{app:prompt_format}.}. We also include a random-selection baseline that samples demonstrations from the full pool. Prompt ordering is kept fixed across settings, and each configuration is repeated over five independent random runs, with results reported as average F1.


\subsection{Results on Beat the AI}

For \textit{Beat the AI}, Figure~\ref{fig:model_size_comparison} compares the behavior of the six models across token budgets, together with the random-selection baseline. A consistent pattern emerges across all model families and sizes: prompts built from the shortest demonstrations (Bin~1) achieve the strongest performance, while longer bins generally underperform as the budget increases. Table~\ref{tab:best_under_4000_compact} summarizes the best F1 for each Bin and also shows the corresponding number of demonstrations under a budget of 4{,}000 tokens. In all six models, the best-performing configuration comes from selecting ICL examples from Bin~1, yielding substantial gains over zero-shot prompting. For instance, Qwen~30B improves by 20.36\% (from 56.38 to 67.86 with $k=51$ total demonstrations), and Gemma~4B improves by 56.20\% (from 38.38 to 59.95 with $k=41$). In contrast, the best configurations from Bins~2--4 are consistently lower, showing that spending the same budget on fewer, longer demonstrations is less effective than allocating it to more concise examples. The random baseline further strengthens this conclusion. Although random selection improves over zero-shot prompting, its performance tends to track medium-length behavior, remaining below the best Bin~1 configurations in every case. This suggests that the gains of Bin~1 are not simply due to using more demonstrations in an arbitrary way, but to deliberately allocating the prompt budget to shorter examples. Overall, the results on \textit{Beat the AI} show that using a larger number of shorter demonstrations is a more effective strategy than relying on fewer longer ones.

\begin{table}[t]
\centering
\scriptsize
\setlength{\tabcolsep}{3pt}
\caption{Best average F1 and its number of demonstrations ($k$) for each length bin under a prompt budget of 4{,}000 tokens on \textit{Beat the AI}. ZS means zero-shot results.}\label{tab:best_under_4000_compact}

\begin{tabular}{@{}l c rr rr rr rr@{}}
\toprule
\multirow{2}{*}{Model} & \multirow{2}{*}{ZS}
& \multicolumn{2}{c}{B1}
& \multicolumn{2}{c}{B2}
& \multicolumn{2}{c}{B3}
& \multicolumn{2}{c}{B4} \\

& 
& \multicolumn{2}{c}{[48--84]}
& \multicolumn{2}{c}{[195--231]}
& \multicolumn{2}{c}{[342--377]}
& \multicolumn{2}{c}{[452--487]} \\
\midrule
Qwen 30B   & 56.38 & \textbf{67.86} & (51) & 64.53 & (18) & 63.34 & (11) & 62.99 & (4) \\
Qwen 4B    & 38.38 & \textbf{59.53} & (56) & 52.19 & (15) & 48.27 & (11) & 46.99 & (7) \\
Phi-4-mini & 43.96 & \textbf{58.01} & (55) & 56.15 & (15) & 55.61 & (10) & 54.60 & (7) \\
Phi-4      & 42.41 & \textbf{66.01} & (51) & 57.48 & (18) & 50.34 & (10) & 48.21 & (8) \\
Gemma 27B  & 55.55 & \textbf{68.52} & (48) & 66.82 & (15) & 65.17 & (11) & 64.00 & (7) \\
Gemma 4B   & 38.38 & \textbf{59.95} & (41) & 58.20 & (15) & 58.11 & (10) & 56.81 & (7) \\
\bottomrule
\end{tabular}
\end{table}

\subsection{Results on Natural Questions}

\begin{table*}[t]
\small
\centering
\setlength{\tabcolsep}{5pt}
\caption{Best average F1 and its number of demonstrations ($k$) for each length bin under a prompt budget of 30{,}000 tokens on NQ. ZS represents zero-shot strategy, and the Random column shows the best average result across five random runs until the token budget is filled.}
\label{tab:nq_best_under_30000}
\begin{tabular}{@{}l c r r r l r l r l r l@{}}
\toprule
\multirow{2}{*}{Model} & \multirow{2}{*}{ZS} 
& \multicolumn{2}{c}{\multirow{2}{*}{Best Random}}
& \multicolumn{2}{c}{B1}
& \multicolumn{2}{c}{B2}
& \multicolumn{2}{c}{B3}
& \multicolumn{2}{c}{B4} \\

& 
& \multicolumn{2}{c}{} 
& \multicolumn{2}{c}{[512--1434]}
& \multicolumn{2}{c}{[2300--3163]}
& \multicolumn{2}{c}{[4029--4892]}
& \multicolumn{2}{c}{[6622--7468]} \\
\midrule
Qwen 30B  & 66.78 & 68.65 & k = 3  & \textbf{70.50} & k = 7  & 68.67 & k = 2 & 66.94 & k = 3 & 69.56 & k = 4 \\
Qwen 4B   & 54.86 & 58.09 & k = 2  & \textbf{61.24} & k = 10 & 60.69 & k = 2 & 59.80 & k = 3 & 55.81 & k = 1 \\
Gemma 4B  & 63.51 & 65.98 & k = 1  & 65.31 & k = 6 & \textbf{66.66} & k = 1 & 65.67 & k = 1 & 63.88 & k = 1 \\
Gemma 27B & 63.24 & 69.83 & k = 6  & \textbf{71.60} & k = 16 & 70.92 & k = 10 & 69.47 & k = 5 & 71.51 & k = 4 \\
\bottomrule
\end{tabular}
\end{table*}

We next evaluate the same prompting strategies on a subset of NQ, using a larger prompt budget of 30{,}000 tokens to reflect the substantially longer contexts in this dataset. Since this setting requires prompts well beyond 16k tokens, we do not include the Phi models in NQ, as Phi-4-mini maximum context length is below the required budget.

Table~\ref{tab:nq_best_under_30000} shows that the pattern on NQ is more nuanced than on \textit{Beat the AI}. Bin~1 yields the best overall result for three of the four evaluated models. The only exception is Gemma~4B, where Bin~2 performs best (66.66 with only one example demonstration, $k=1$), although the gap with Bin~1 is small. Thus, shorter demonstrations remain highly competitive, but the preference for Bin~1 is less uniform in this long-context setting.

\begin{figure*}[t]
    \centering

    \captionsetup[subfigure]{skip=2pt}
    
    \begin{subfigure}[t]{0.49\textwidth}
        \centering
        \includegraphics[width=\linewidth]{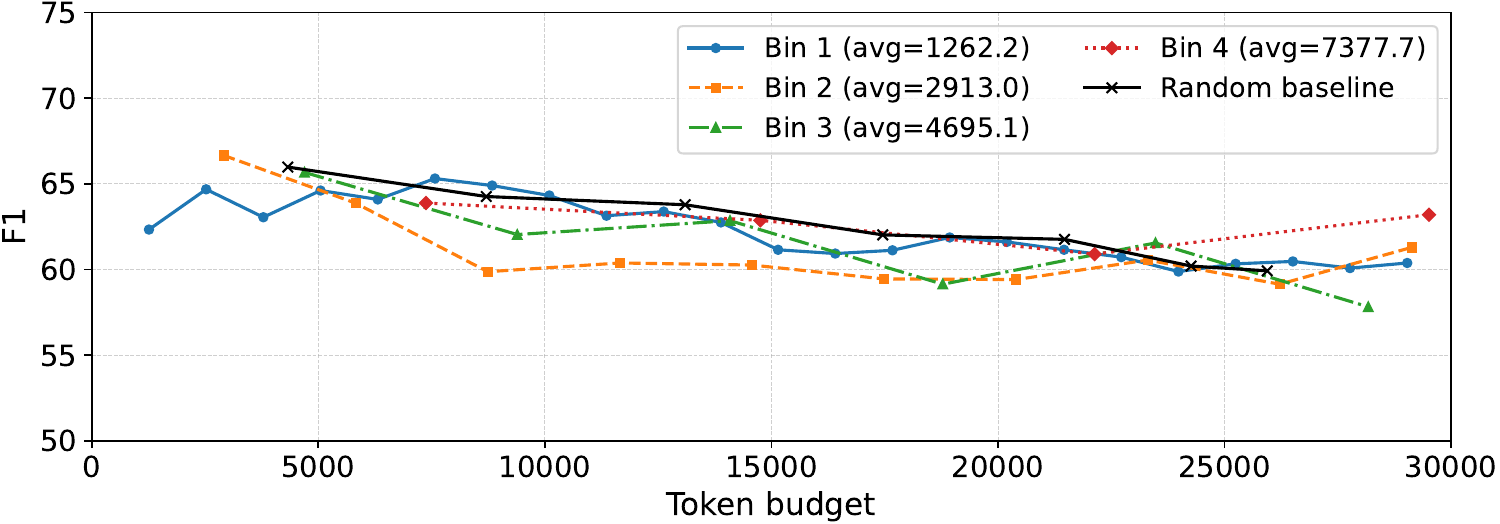}
        \caption{Gemma 4B}
        \label{fig:gemma4b}
    \end{subfigure}\hfill
    \begin{subfigure}[t]{0.49\textwidth}
        \centering
        \includegraphics[width=\linewidth]{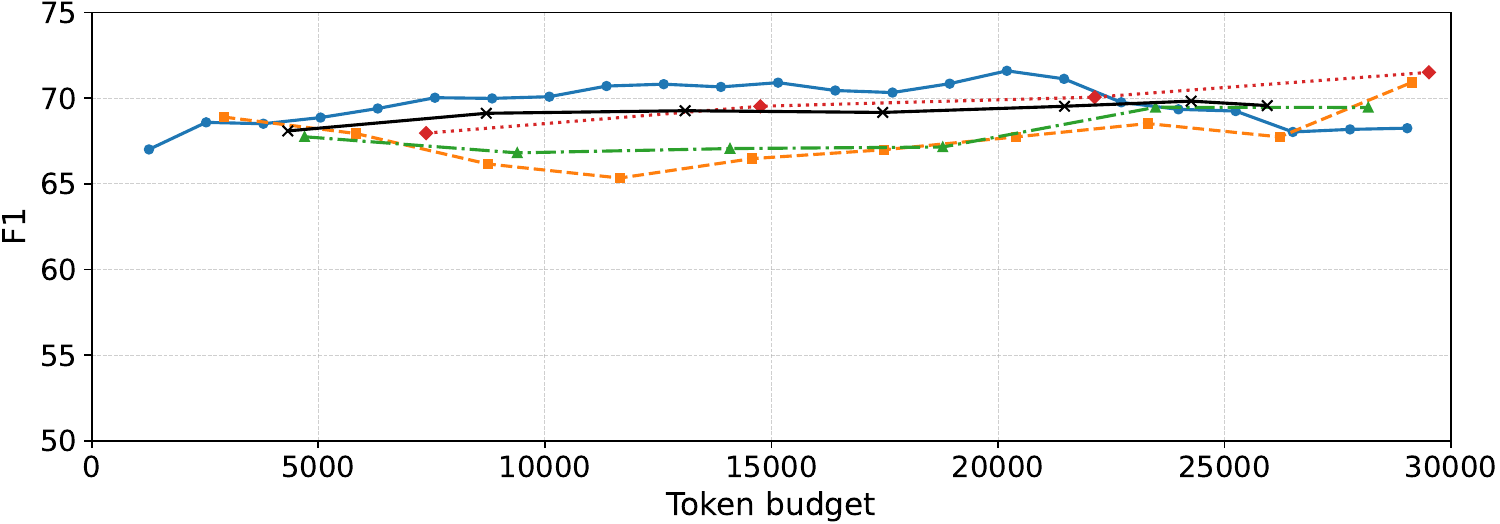}
        \caption{Gemma 27B}
        \label{fig:gemma27b}
    \end{subfigure}

    \vspace{-0.2em}

    \begin{subfigure}[t]{0.49\textwidth}
        \centering
        \includegraphics[width=\linewidth]{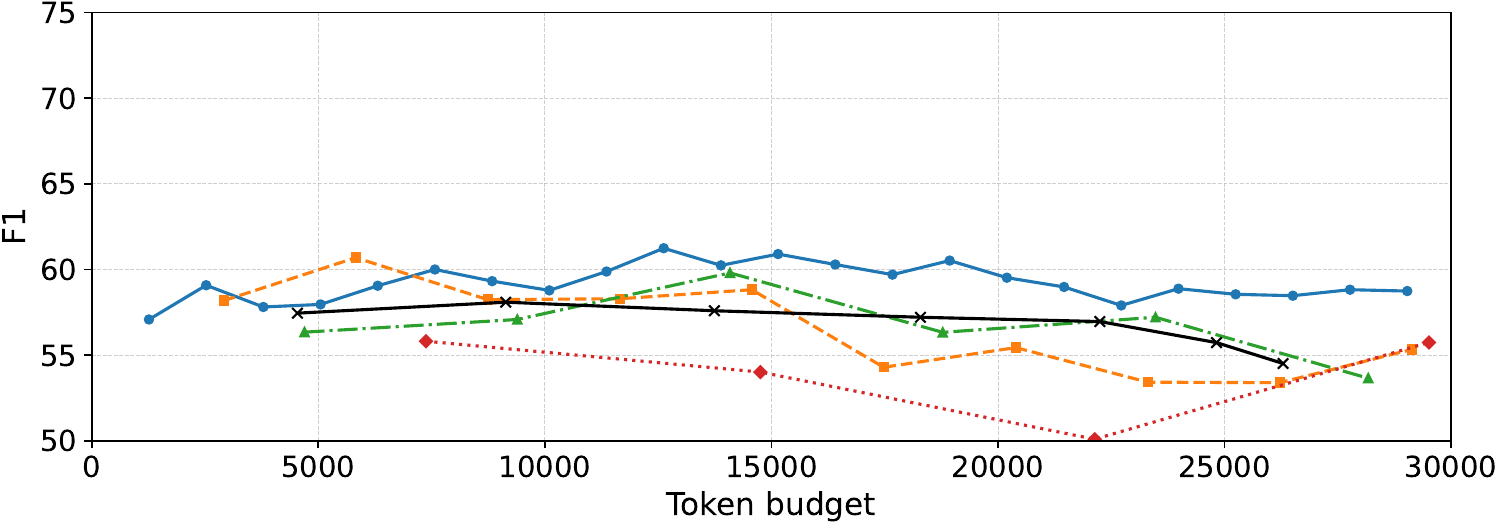}
        \caption{Qwen 4B}
        \label{fig:qwen4b}
    \end{subfigure}\hfill
    \begin{subfigure}[t]{0.49\textwidth}
        \centering
        \includegraphics[width=\linewidth]{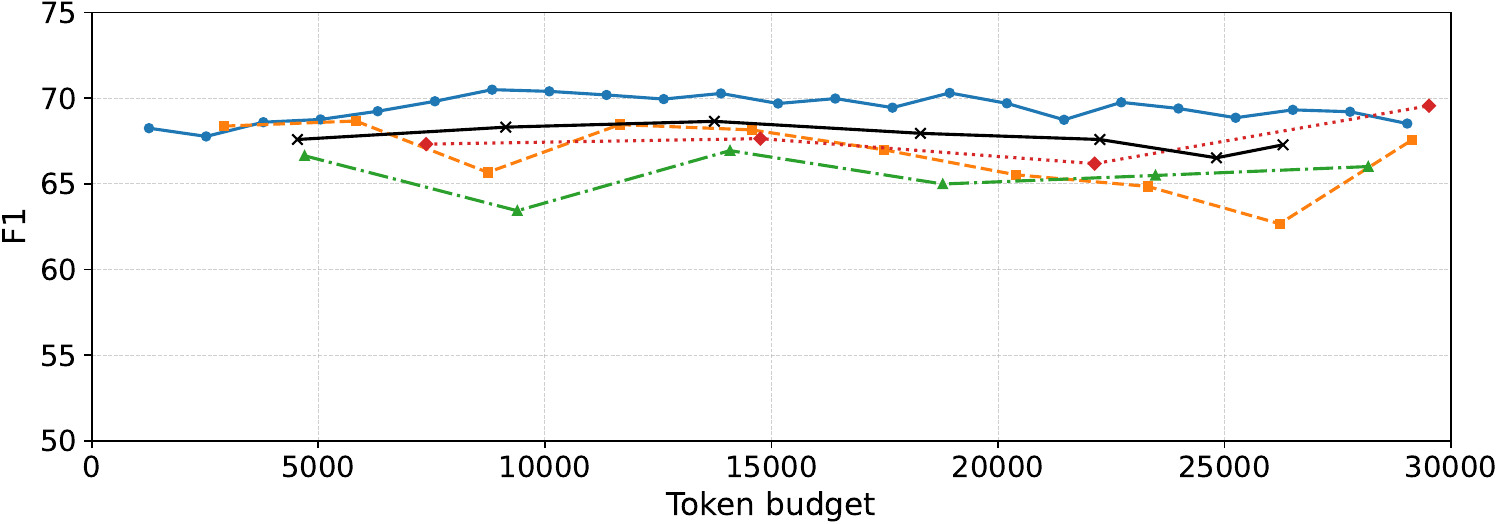}
        \caption{Qwen 30B}
        \label{fig:qwen30b}
    \end{subfigure}

    \vspace{-0.4em}
\caption{Average F1 on \textit{NQ} as a function of prompt budget for the four models under different length bins. Each curve corresponds to a length bin, and random baseline samples demonstrations from the full pool.}
    \label{fig:nq_model_size_comparison}
\end{figure*}

Figure~\ref{fig:nq_model_size_comparison} complements these results by showing performance across prompt budgets. Increasing the budget does not consistently improve F1: Gemma~4B generally deteriorates as prompts grow longer, while the other models exhibit plateaus or fluctuations. Bin~1 remains competitive across much of the budget range, particularly for the Qwen models, although its advantage is not uniform.

All models improve over ZS prompting, but the margins between the best bin configurations are narrower than in \textit{Beat the AI}, and the best configurations are often reached with small values of $k$. The random baseline supports this picture: for all four models, its best result remains below the strongest bin-based configuration and is achieved with only a small number of demonstrations, further indicating that filling the available budget does not necessarily yield the best performance.

\subsection{Discussion of QA Results}

Taken together, the results show that the benefit of shorter demonstrations is robust, but not equally strong across QA settings. On \textit{Beat the AI}, the effect is clear and consistent: the shortest bin yields the best results for all evaluated models, often by a substantial margin. This indicates that, when demonstrations are relatively compact, allocating a fixed token budget to more examples is a particularly effective strategy.

On NQ, the same preference for shorter demonstrations remains, but the pattern is weaker and less uniform. The differences between bins are smaller, and the best configurations are often reached with relatively small values of $k$. Because NQ relies on full Wikipedia articles, even the shortest demonstrations already consume substantial context space, so adding examples quickly produces longer prompts. From an attention perspective, this creates two related problems. First, the model’s finite attention must be distributed over many more tokens, so each additional demonstration receives less effective focus. Second, longer prompts make it harder for the model to recover relevant information, especially when useful evidence is placed in the middle of the sequence. As a result, increasing $k$ does not necessarily improve performance: the potential benefit of extra demonstrations is offset by weaker attention and more difficult evidence retrieval. This interpretation is consistent with prior work on long-context limitations, including the ``lost in the middle'' phenomenon~\cite{liu2023lost,zou2025many,Gabin-SIGIR-RAG-Lost}.

\subsection{Cross-Task Experiment on ANLI R3}
\label{sec:anli_r3}

To explore whether the patterns observed in extractive QA extend to another task, we conduct an additional experiment on natural language inference (NLI), using the third round of the Adversarial NLI benchmark (ANLI R3)~\cite{nie-etal-2020-adversarial}. This task requires classifying the relation between a premise and a hypothesis as entailment, contradiction, or neutral. We evaluate the same six models using four demonstration-length bins and report macro-F1. Table~\ref{tab:anli_best_under_4000} presents the best observed configuration within each bin under a full-prompt token budget of 4{,}000 tokens.

\begin{table}[t]
\centering
\scriptsize
\setlength{\tabcolsep}{3pt}
\caption{Best observed macro-F1 and number of demonstrations ($k$, in parentheses) per length bin on ANLI R3. Bold marks the best bin score per model.}
\label{tab:anli_best_under_4000}

\begin{tabular}{@{}l c rr rr rr rr@{}}
\toprule
\multirow{2}{*}{Model}
& \multirow{2}{*}{ZS}
& \multicolumn{2}{c}{B1}
& \multicolumn{2}{c}{B2}
& \multicolumn{2}{c}{B3}
& \multicolumn{2}{c}{B4} \\
&
& \multicolumn{2}{c}{[0--63]}
& \multicolumn{2}{c}{[100--118]}
& \multicolumn{2}{c}{[155--173]}
& \multicolumn{2}{c}{[210--411]} \\
\midrule
Qwen 30B   & 60.5 & 62.06 & (2)  & \textbf{64.09} & (3)  & 62.34 & (1)  & 62.75 & (4)  \\
Qwen 4B    & 47.0 & 50.82 & (67) & \textbf{53.15} & (28) & 49.56 & (24) & 52.22 & (18) \\
Phi-4-mini & 45.5 & 49.41 & (6)  & 48.95 & (29) & 49.23 & (8)  & \textbf{50.38} & (1)  \\
Phi-4      & 61.5 & \textbf{63.85} & (6) & 62.59 & (28) & 62.30 & (1) & 63.31 & (17) \\
Gemma 27B  & 60.0 & 62.90 & (48) & \textbf{66.84} & (24) & 64.16 & (6) & 66.09 & (14) \\
Gemma 4B   & 46.5 & \textbf{51.93} & (55) & 51.41 & (24) & 48.84 & (5) & 50.03 & (13) \\
\bottomrule
\end{tabular}
\end{table}

The two shorter bins yield the highest observed macro-F1 for five of the six models. Bin~1 performs best for Phi-4 and Gemma~4B, while Bin~2 leads for both Qwen models and Gemma~27B. Phi-4-mini achieves its best result with a single Bin~4 demonstration. Longer demonstrations also remain competitive: for Phi-4 and Gemma~27B, the best Bin~4 score is within one macro-F1 point of the overall best configuration. Across models, the winning configurations improve over zero-shot prompting by 2.35--6.84 macro-F1 points, with demonstration counts ranging from $k=1$ to $k=55$.

These results show that the preferred allocation of the token budget depends on the task and model. Compared with the consistent Bin~1 advantage on \textit{Beat the AI}, ANLI favors a broader range of demonstration lengths and counts. Its bins also exhibit less separation in token length than those used in the QA experiments, providing a complementary setting in which longer demonstrations remain competitive. Investigating how the range of available demonstration lengths interacts with task demands is a promising direction for future work.

\section{Conclusions}

We studied the trade-off between demonstration length and demonstration count in QA-oriented ICL under fixed token budgets. Across both QA datasets, shorter demonstrations are generally the most reliable use of prompt space. This effect is especially strong on \textit{Beat the AI}, while on NQ it remains visible but weaker, as long contexts quickly consume the budget. An additional experiment on ANLI R3 extends the analysis beyond extractive QA: the two shorter bins provide the best observed configurations for five of six models, although longer demonstrations remain competitive. Overall, our findings highlight demonstration granularity as an important factor in prompt design and suggest that shorter demonstrations are usually a better investment under token constraints, with the preferred length and count depending on the task and model.

\section{Computational Resources}

Experiments were conducted using private infrastructure, with a carbon efficiency of 0.432 kgCO$_2$eq/kWh. A cumulative of 53 hours of computation was performed on hardware of type A100 SXM4 80 GB, generating an estimated 7.5 kgCO$_2$eq \cite{lacoste2019quantifying}. 


\section*{Acknowledgment}

This work was supported by grant PID2022-137061OB-C21 and PID2025-167749OB-C22, funded by MICIU/AEI/10.13039/501100011033 and ERDF/EU; and the Consellería de Cultura, Educación, Formación Profesional e Universidades through accreditations ED431G 2023/01 and ED431C 2025/49. CITIC, a CIGUS Network member accredited for excellence within the Galician University System, receives support from the Department of Education, Science, Universities, and Vocational Training of the Xunta de Galicia and is co-financed by the EU through the FEDER Galicia 2021--2027 operational programme (Ref.~ED431G 2023/01).

\section*{Limitations}

This study has several limitations. First, it focuses primarily on QA-oriented in-context learning, with an additional experiment on natural language inference using ANLI R3. Although this extends the evaluation beyond QA, the findings may not generalize directly to other tasks. Second, although we evaluate multiple LLM families and model sizes, the experiments are still limited to a finite set of instruction-tuned models. Finally, our setup uses prompts built from a single length bin at a time and random sampling within each bin, which enables comparisons across demonstration lengths but does not cover mixed-bin strategies or relevance-based selection.

\section*{Ethical use of AI-assisted tools}
AI-assisted tools were used solely to support manuscript review and language editing. All suggestions were critically reviewed and revised by the authors, who retained full control over the manuscript and its scientific content. 

\bibliography{custom}

\appendix

\section{Prompt Format}
\label{app:prompt_format}

\subsection{Question Answering}

All QA experiments use the same task instruction and multi-turn prompt format across datasets. Each prompt begins with a single system instruction defining the extraction task. The model is then given a sequence of in-context demonstrations, where each demonstration contains a question, its associated context, and the expected answer in the required output format. Finally, the test instance is appended using the same structure, without the gold answer.

The common task instruction used in both \textit{Beat the AI} and Natural Questions is shown in Table~\ref{tab:task_prompt}.

\begin{table*}[t]
\centering
\small
\setlength{\fboxsep}{6pt}
\fbox{%
\begin{minipage}{0.96\textwidth}
\ttfamily
You are an information extraction system.

You must answer the QUESTION using ONLY the CONTEXT.

Task: \\
- If the CONTEXT contains an answer, extract it as a contiguous span copied verbatim from the CONTEXT. \\
- If the CONTEXT does not contain enough information to answer, output NO\_ANSWER.

Rules: \\
1) The extracted ANSWER must be an EXACT contiguous substring from CONTEXT (verbatim, case-sensitive). \\
2) Prefer the minimal span that answers the question precisely and unambiguously. \\
3) If no such span exists in CONTEXT, output exactly: NO\_ANSWER \\
4) Output exactly one line with this format and nothing else: \\
\hspace*{1.5em}ANSWER: <answer span or NO\_ANSWER>
\end{minipage}}
\caption{Common task instruction used across both QA datasets and models.}
\label{tab:task_prompt}
\end{table*}

Each in-context demonstration follows the same multi-turn structure:

\begin{center}
\fbox{%
\begin{minipage}{0.92\columnwidth}
\small\ttfamily
QUESTION: \\[2pt]
<question text>

CONTEXT: \\[2pt]
<context text>

ANSWER: \\[2pt]
<gold answer span or NO\_ANSWER>
\end{minipage}}
\end{center}

The final test instance is formatted identically, but without the gold answer:

\begin{center}
\fbox{%
\begin{minipage}{0.92\columnwidth}
\small\ttfamily
QUESTION: \\[2pt]
<test question>

CONTEXT: \\[2pt]
<test context>

ANSWER:
\end{minipage}}
\end{center}

This format ensures that all models receive the same extraction instruction and answer constraints, while varying only the number and length of the demonstrations included under each token budget.

\subsection{Natural Language Inference}
\label{app:nli_prompt_format}

For ANLI R3, we use a classification instruction and a multi-turn prompt format. Each prompt begins with the system instruction shown in Table~\ref{tab:nli_task_prompt}. Each in-context demonstration consists of a user message containing a premise, introduced as \texttt{CONTEXT}, and a hypothesis, followed by an assistant message containing the gold label. The final test instance is appended as a user message with the same structure, without its gold label.

\begin{table*}[t]
\centering
\small
\setlength{\fboxsep}{6pt}
\fbox{%
\begin{minipage}{0.96\textwidth}
\ttfamily
You are a Natural Language Inference (NLI) classification system.

You must decide the logical relationship between a CONTEXT (premise) and a HYPOTHESIS.

Task: \\
- Read the CONTEXT and the HYPOTHESIS carefully. \\
- Output exactly one label: \\
\hspace*{1em}* Entailment --- the CONTEXT clearly supports / implies the HYPOTHESIS \\
\hspace*{1em}* Neutral --- the CONTEXT neither supports nor contradicts the HYPOTHESIS \\
\hspace*{1em}* Contradiction --- the CONTEXT clearly contradicts the HYPOTHESIS

Rules: \\
1) Output exactly one line with this format and nothing else: \\
LABEL: <Entailment | Neutral | Contradiction> \\
2) Do not add any explanation or extra text.
\end{minipage}}
\caption{System instruction used for ANLI R3 across all evaluated models.}
\label{tab:nli_task_prompt}
\end{table*}

Each demonstration is represented by the following pair of messages.

\noindent\textbf{User message:}
\begin{center}
\fbox{%
\begin{minipage}{0.92\columnwidth}
\small\ttfamily
CONTEXT: \\[2pt]
<premise text>

HYPOTHESIS: \\[2pt]
<hypothesis text>
\end{minipage}}
\end{center}

\noindent\textbf{Assistant message:}
\begin{center}
\fbox{%
\begin{minipage}{0.92\columnwidth}
\small\ttfamily
LABEL: <gold label>
\end{minipage}}
\end{center}

The gold label is one of \texttt{Entailment}, \texttt{Neutral}, or \texttt{Contradiction}. The final user message contains only the test premise and hypothesis:

\begin{center}
\fbox{%
\begin{minipage}{0.92\columnwidth}
\small\ttfamily
CONTEXT: \\[2pt]
<test premise>

HYPOTHESIS: \\[2pt]
<test hypothesis>
\end{minipage}}
\end{center}

The model then generates the classification as an assistant response in the required \texttt{LABEL: <label>} format. The system instruction and message structure remain unchanged across models and demonstration-length bins.

\section{Length-Bin Construction and Selection}
\label{app:bins}

To construct the demonstration-length bins used in our experiments, we first analyzed the token-length distribution of the candidate training examples in each dataset using histogram partitions based on the Freedman--Diaconis rule~\cite{freedman1981histogram}. We then defined four representative ranges for each dataset, aiming to balance two criteria: (i) sufficient numbers of examples in each range, and (ii) clear separation in token length between ranges. The histogram partitions describe the underlying distributions, while the four experimental bins define the candidate pools used for demonstration sampling.

\subsection{Beat the AI}

\begin{figure*}[htbp]
    \centering
    \includegraphics[width=0.9\textwidth]{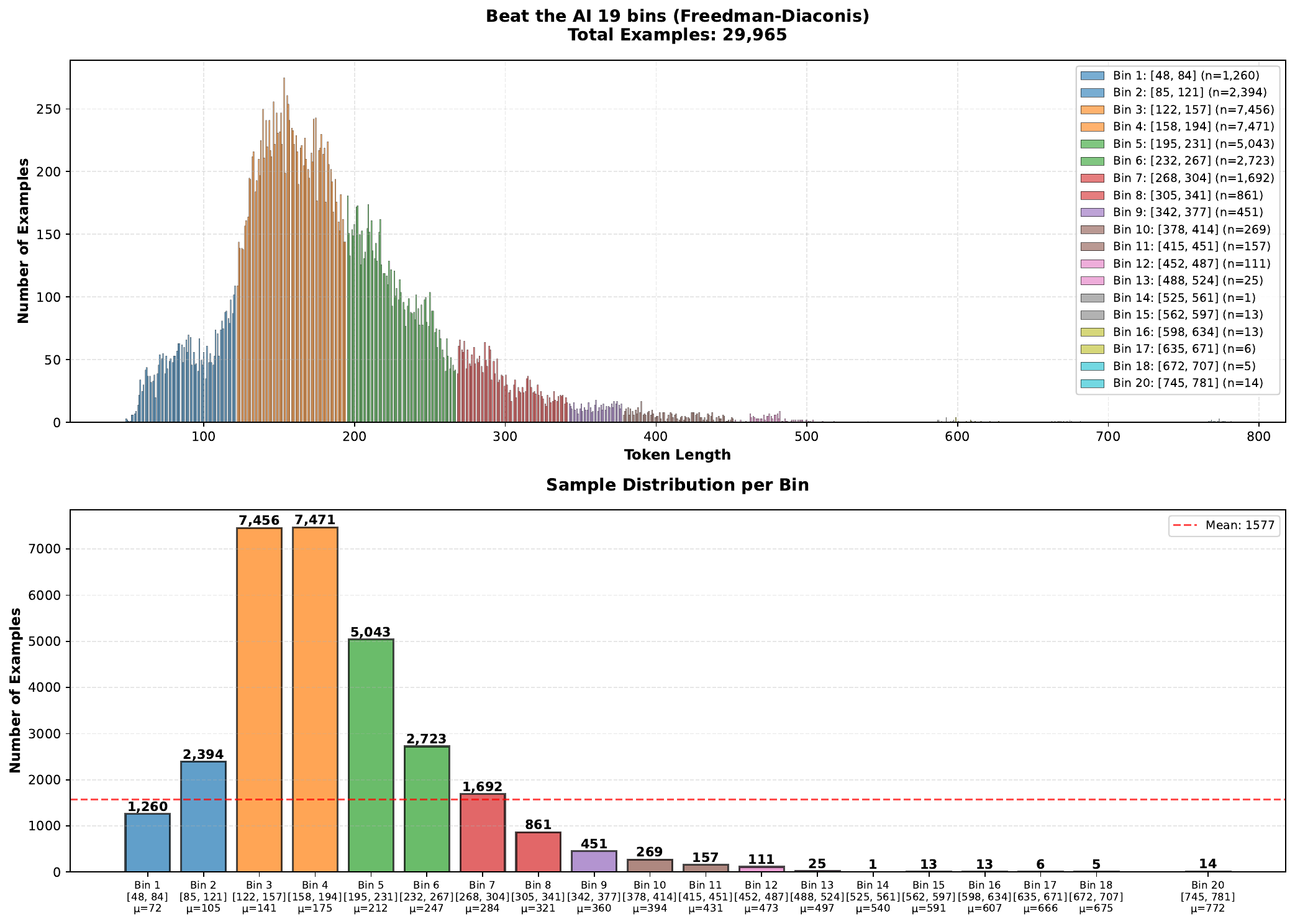}
    \caption{Token-length distribution and Freedman--Diaconis bins for the \textit{Beat the AI} candidate examples. The top plot shows the token-length histogram, and the bottom plot shows the number of examples per histogram bin.}
    \label{fig:appendix_squad_bins}
\end{figure*}

For the \textit{Beat the AI} setting, the Freedman--Diaconis procedure produced 19 bins over 29,965 examples, spanning token lengths from 48 to 781. As shown in Figure~\ref{fig:appendix_squad_bins}, the resulting distribution is concentrated in the short and medium ranges, with progressively fewer examples in the longer bins. Based on this distribution, we selected four bins for the experiments: [48--84], [195--231], [342--377], and [452--487]. These bins provide a reasonable trade-off between coverage and length separation.

\subsection{Natural Questions}

\begin{figure*}[htbp]
    \centering
    \includegraphics[width=0.9\textwidth]{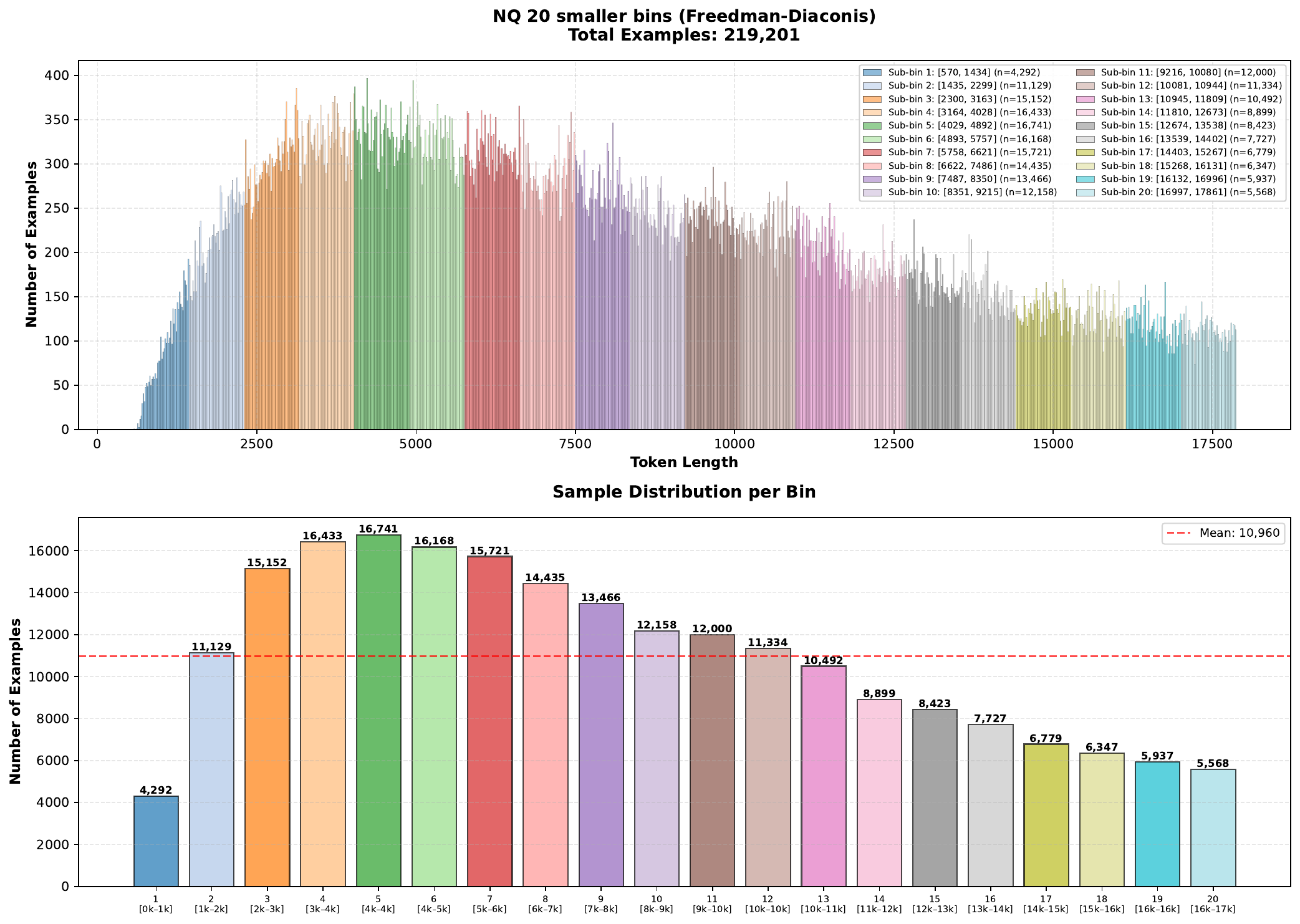}
    \caption{Token-length distribution and histogram bins for the NQ candidate examples. The top plot shows the token-length histogram, and the bottom plot shows the number of examples per histogram bin.}
    \label{fig:appendix_nq_bins}
\end{figure*}

For Natural Questions, we followed the same procedure on the subset of candidate examples shorter than 18{,}000 tokens, obtaining 20 smaller bins over 219,201 examples. As shown in Figure~\ref{fig:appendix_nq_bins}, the distribution is much broader than in \textit{Beat the AI}, reflecting the substantially longer contexts in NQ. From these bins, we selected four representative ranges for the experiments: [512--1434], [2300--3163], [4029--4892], and [6622--7468]. Compared with \textit{Beat the AI}, these bins are both larger and farther apart, which is consistent with the heavier contextual structure of NQ examples.

\subsection{ANLI R3}

\begin{figure*}[htbp]
    \centering
    \includegraphics[width=0.9\textwidth]{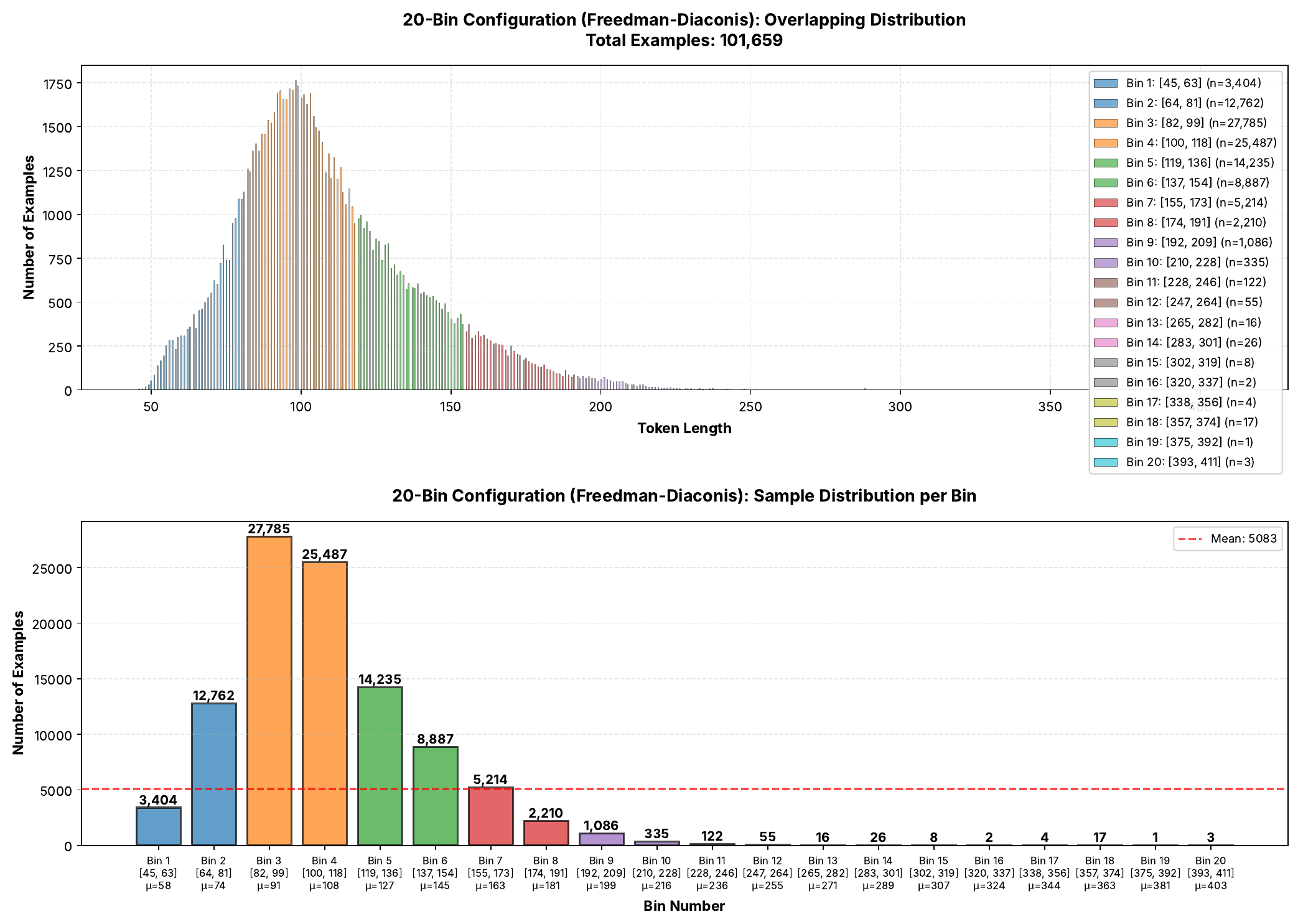}
    \caption{Token-length distribution of the ANLI R3 candidate examples, displayed using 20 histogram bins. The top plot shows the token-length histogram, and the bottom plot reports the number of examples and mean token length ($\mu$) per histogram bin. Experimental bins $B_1$, $B_2$, and $B_3$ correspond to histogram bins 1, 4, and 7, respectively; $B_4$ combines histogram bins 10--20.}
    \label{fig:appendix_anli_bins}
\end{figure*}

For ANLI R3, Figure~\ref{fig:appendix_anli_bins} shows a 20-bin histogram over 101,659 candidate examples, spanning token lengths from 45 to 411. The distribution is concentrated around 82--118 tokens and exhibits a sparse tail toward longer examples. The two most populated histogram intervals, [82--99] and [100--118], contain 27,785 and 25,487 examples, respectively, whereas examples longer than 209 tokens account for fewer than 1\% of the candidate pool.

Based on this distribution, we constructed four experimental bins: $B_1=[0,63]$, $B_2=[100,118]$, $B_3=[155,173]$, and $B_4=[210,411]$. The first three correspond to histogram bins 1, 4, and 7, containing 3,404, 25,487, and 5,214 examples, respectively. Although the lower bound of $B_1$ is set to zero, its observed examples span 45--63 tokens. For $B_4$, we pooled histogram bins 10--20, including all examples from 210 tokens to the maximum observed length of 411 tokens, yielding 589 candidates. Pooling these sparse intervals provides a larger candidate pool for the longest demonstrations while preserving clear length separation from the other experimental bins.

\end{document}